\documentclass[runningheads]{llncs}

\usepackage{times}
\usepackage{latexsym}
\usepackage{tabularx}
\usepackage{graphicx}
\usepackage{csquotes}
\usepackage{booktabs}
\usepackage{multicol}
\usepackage{url}
\usepackage{xspace}
\usepackage{amsmath}
\usepackage{amsfonts}
\usepackage[colorlinks, linkcolor=blue, urlcolor=blue]{hyperref}
\usepackage{algpseudocode}
\usepackage{algorithm}
\usepackage{algorithmicx}
\usepackage{breakcites}

\renewcommand{\listtablename}{}
\renewcommand{\listfigurename}{}

\newcommand{\ind}{\hspace{\algorithmicindent}}

\begin{document}

\title{Call Larisa Ivanovna: Code-Switching Fools Multilingual NLU Models}

\titlerunning{Code-Switching Fools Multilingual NLU Models}
\author{Alexey Birshert  \and Ekaterina Artemova  }

\authorrunning{Birshert and Artemova}

\institute{National Research University Higher School of Economics, Moscow, Russia
\url{https://cs.hse.ru/en/ai/computational-pragmatics/} 
\email{\\ \{abirshert,elartemova\}@hse.ru} }

\maketitle
\begin{abstract}
Practical needs of developing task-oriented dialogue assistants require the ability to understand many languages. Novel benchmarks for multilingual natural language understanding (NLU) include monolingual sentences in several languages, annotated with intents and slots. In such setup models for cross-lingual transfer show remarkable performance in joint intent recognition and slot filling. However, existing benchmarks lack of code-switched utterances, which are difficult to gather and label due to complexity in the grammatical structure. The evaluation of NLU models seems biased and limited, since code-switching is being left out of scope. 

Our work adopts recognized methods to generate plausible and naturally-sounding code-switched utterances and uses them to create a synthetic code-switched test set. Based on experiments, we report that the state-of-the-art NLU models are unable to handle code-switching. At worst, the performance, evaluated by semantic accuracy, drops as low as 15\% from 80\% across languages. Further we show, that pre-training on synthetic code-mixed data helps to maintain performance on the proposed test set at a comparable level with monolingual data. Finally, we analyze different language pairs and show that the closer the languages are, the better the NLU model handles their alternation. This is in line with the common understanding of how multilingual models conduct transferring  between languages. 

\end{abstract}

\section{Introduction} The usability of task-oriented dialog (ToD) assistants depends crucially on their ability to process users' utterances in many languages. At the core of a task-oriented dialog assistant is a natural language understanding (NLU) component, which parses an input utterances into a semantic frame by means of intent recognition and slot filling \cite{tur2010left}. Intent recognition tools identify user needs (such as \texttt{buy a flight ticket}). Slot filling extracts the intent's arguments (such as \texttt{departure city} and \texttt{time}). 

Common approaches to training multilingual task-oriented dialogue systems rely on (i) the abilities of pre-trained language models to transfer learning across languages and \cite{conneau2019unsupervised,liu2020multilingual} and (ii) translate-and-align pipelines \cite{hu2020xtreme}.

The latter group of methods leverages upon pre-training on large amounts of raw textual data for many languages. Different learning objectives are used to train representations aligned across languages. The alignment can be further improved by means of task-specific label projection  \cite{liu2020cross} and representation alignment \cite{gritta2021xeroalign} methods. 

The former group of methods utilizes off-the-shelf machine translation engines to translate (i) the training data from resource-rich languages (almost exclusively English) to target languages or (ii) the evaluation data from target languages into English \cite{hu2020xtreme}. Further word alignment techniques help to match slot-level annotations \cite{einolghozati2021volumen,nicosia2021translate}. Finally, a monolingual model is trained to make desired predictions. 

A number of datasets for cross-lingual NLU have been developed. To name a few, MultiAtis++ \cite{xu2020end}, covering seven languages across four language families, contains dialogues related to a single domain -- air travels. MTOP \cite{li2021mtop} comprises six languages and 11 domains.  xSID \cite{vanmasked} is an evaluation-only small-scale dataset, collected for 12 languages and six domains.

Recent research has adopted a new experimental direction aimed at developing cross-lingual augmentation techniques, which learn the inter-lingual semantics across languages \cite{einolghozati2021volumen,guo2021learning,krishnan2021multilingual,liu2020attention,qin2020cosda}. These work seek to simulate code-switching, a phenomenon where speakers use multiple languages mixed up \cite{sankoff1981formal}. Experimental results consistently show that augmentation with synthetic code-switched data leads to significantly improved performance for cross-lingual NLU tasks. What is more, leveraging upon these datasets in practice meets the needs of multicultural and multilingual communities. To the best of our knowledge, large-scale code-switching ToD corpora do not exist. 

This paper extends the ongoing research on exploring benefits from synthetic code-switching to cross-lingual NLU. Our approach to generating code-switched utterances relies on grey-box adversarial attacks on the NLU model. We perturb the source utterances by replacing words or phrases with their translations to another language. Next, perturbed utterances are fed to the NLU model. Increases in the loss function indicate difficulties in making predictions for the utterance. This way, we can (i) generate code-switched adversarial utterances, (ii) discover insights on how code-switching with different languages impacts the performance of the target language, (iii) gather augmentation data for further fine-tuning of the language model. 
To sum, our contributions are\footnote{Our code can be found at \url{https://github.com/PragmaticsLab/CodeSwitchingAdversarial}}:
\begin{enumerate}
    \item We implement several simple heuristics to generate  code-switched utterances based on monolingual data from an NLU benchmark;
    \item We show-case that monolingual models fail to process code-switched utterances. At the same time, cross-lingual models cope much better with such texts;
    \item We show that fine-tuning of the language model on code-switched utterances improves the overall semantic parsing performance by up to a 2-fold increase. 
\end{enumerate}

\section{Related work}

\subsubsection{Generation of code-switched text} has been explored as a standalone task \cite{gupta2020semi,lee2020modeling,rizvi2021gcm,samanta2019deep,tan2021code,winata2019code} and as a way to augment training data for cross-lingual applications, including task-oriented dialog systems  \cite{einolghozati2021volumen,guo2021learning,krishnan2021multilingual,liu2020attention,qin2020cosda}, machine translation \cite{abdul2021exploring,gautam2021comet,gupta2021training}, natural language inference and question answering \cite{singh2019xlda}, speech recognition \cite{yilmaz2018acoustic}.

Methods for generating code-switched text range from simplistic re-writing of some words in the target script \cite{gautam2021comet} to adversarial attacks on cross-lingual pre-trained language models \cite{tan2021code} and building complex hierarchical VAE-based (Variational AutoEncoders) models \cite{samanta2019deep}.  The vast majority of methods utilize machine translation engines \cite{singh2019xlda}, parallel datasets \cite{abdul2021exploring,gautam2021comet,gupta2021training,winata2019code} or bilingual lexicons \cite{tan2021code}  to replace the segment of the input text with its translations. Bilingual lexicons may be induced from the parallel corpus with the help of soft alignment, produced by attention mechanisms \cite{lee2020modeling,liu2020attention}. Pointer networks can be used to select segments for further replacement \cite{gupta2020semi,winata2019code}.  If natural code-switched data is available, such segments can be identified with a sequence labeling model \cite{gupta2021training}. Other methods rely on linguistic theories of code-switching. To this end, GCM toolkit \cite{rizvi2021gcm} leverages two linguistic theories, which help to identify segments where code-switching may occur by aligning words and mapping parse trees of parallel sentences. 

The quality of generated code-switched texts is evaluated by (i) {\bf intrinsic} text properties, such as code-switching ratio, length distribution, and (ii) {\bf extrinsic} measures, ranging from the perplexity of external languages model to the downstream task performance, in which code-switched data was used for augmentation \cite{samanta2019deep}.  

\subsubsection{Natural Language Understanding} in the ToD domain has two main goals, namely intent recognition, and slot filling \cite{razumovskaia2021crossing}. Intent recognition assigns a user utterance with one of the pre-defined intent labels. Thus, intent recognition is usually tackled with classification methods. Slot filling seeks to find arguments of the assigned intent and is modeled as a sequence modeling problem. For example, the utterance {\it I need a flight from Moscow to Tel Aviv on 2nd of December} should be assigned with the intent label \texttt{find flight}; three slots may be filled: \texttt{departure city}, \texttt{arrival city}, \texttt{date}.  These two interdependent NLU tasks are frequently approached via multitask learning, in which a joint model is trained to recognize intents and fill in slots simultaneously  \cite{chen2019bert,DBLP:conf/acl/KimSKK18,wu-etal-2020-tod}.

\subsubsection{Adversarial attacks} on natural language models has been categorized with respect to (i) what kind of information is provided from the model and (ii) what kind of perturbation is applied to the input text \cite{morris2020textattack}. White-box attacks \cite{ebrahimi2018hotflip} have access to the whole model's inner workings. On the opposite side, black-box attacks \cite{gao2018black} do not have any knowledge about the model. Grey-box attacks \cite{xu2021grey} access predicted probabilities and loss function values. Perturbations can be applied at char-, token-, and sentence-levels \cite{gao2018black,li2020bert,cheng2020seq2sick}. 

\subsubsection{Other related research directions} include {\bf code-switching detection} \cite{mave-etal-2018-language,sravani-etal-2021-political}, evaluation of pre-trained language models' {\bf robustness} to code-switching \cite{winata2021multilingual}, analysis of language model's inner workings with respect to code-switched inputs \cite{santy2021bertologicomix}, \textbf{benchmarking} downstream tasks in code-switched data   \cite{aguilar-etal-2020-lince,khanuja2020gluecos}.

\section{Our approach} In our work we train multilingual language models for the joint intent recognition and slot-filling task.

\subsection{Dataset} We chose MultiAtis++ dataset  \cite{xu2020end} as the main source of data. This dataset contains seven languages from three language families - Indo-European (English, German, Portuguese, Spanish, and French), Japanese, and Sino-Tibetan (Chinese). The dataset is a parallel corpus for classifying intents and slot labeling - in 2020 it was translated from English to the other six languages. The training set contains 4978 samples for each language; the test set contains 893 samples per language. Each object in the dataset consists of a sentence, slot labels in BIO format, and the intent label.

\subsection{Joint intent recognition and slot-filling} We train a single model for the joint intent recognition and slot-filling task. The model has two heads; the first one predicts intents, and the second one predicts slots. We had trained two different models as a backbone - m-BERT and XLM-RoBERTa.

We aim at comparing two setups: (i) training on the whole dataset and (ii) only on its English subset followed by zero-shot inference for other languages. For convenience, we propose short names for the four models trained during the research - \textbf{xlm-r, xlm-r-en, m-bert, m-bert-en}.

We measure our models' quality with three metrics: \textbf{intent accuracy}, \textbf{F1 score} for slots (we used micro-averaging by classes) and the proportion of sentences where we correctly classified everything - both intent and all slots - \textbf{semantic accuracy}.

\subsection{Code-switching generation} We propose two variants of gray-box adversarial attacks. During the attack, we have access to the model's loss of input data. We strive to create an attack so that the resulting adversarial perturbation of the source sentence is as close as possible to the realistic code-switching. Quality evaluation at such adversarial attacks can act as a lower bound for corresponding models' quality in similar problems in the presence of real code-switching in input data.
\begin{algorithm}[!h]
    \caption{General view of the attack}
    \begin{algorithmic}
        \Require{Sentence and label x, y; source model $\mathcal{M}$; embedded target language $L$}
        \Ensure{Adversarial sample x'} \\
        $\mathcal{L}_{x}$ = GetLoss($\mathcal{M}$, x, y)
        \For{i in permutation(len(x))}
            \\
            \ind Candidates = GetCandidates($\mathcal{M}$, x, y, token\_id = i) \\
            \ind Losses = GetLoss($\mathcal{M}$, Candidates)
            \ind\If{Candidates and max(Losses) $\geq$ $\mathcal{L}_{x}$}
                    \\
                    \ind\ind$\mathcal{L}_{x}$ = max(Losses) \\
                    \ind\ind x, y = Candidates[argmax(Losses)]
            \EndIf
        \EndFor \\
        \ind\Return x
    \end{algorithmic}\label{alg:algorithm}
\end{algorithm}
We focus mainly on the lexical aspect of code-switching when some words are replaced with their substitutes from other languages. We replace some tokens in the source sentence with their equivalents from the attacking languages during the attack. The method to determine the replacement depends on which exactly attack is used. Since most people who can use code-switching are bilinguals, in our work, we propose to analyze attacks consisting in embedding one language into another.

\subsubsection{Overview of the attacks} The general attack scheme (Algorithm~\ref{alg:algorithm}) is the same for both proposed attacks. We offer the following attack pattern in our work: a source model, a pair of sample sentences and labels, and embedded target language. Then we iterate over the tokens in the sample sentence and strive to replace them with their equivalent from the embedded target language. If changing the token to its equivalent increases the source model's loss, we replace the token with the proposed candidate. The difference between the two methods consists in the way they generate replacement candidates.

\subsubsection{Word-level adversaries} The first attack (Algorithm~\ref{alg:algorithm1}) generates target embedded language substitutions by translating single tokens into the corresponding language. Attacking this way, we make a rough lower bound since we do not consider the context of the sentences and the ambiguity of words during the attack. To translate words into other languages, we use the large-scale many-to-many machine translation model M2M-100 from Facebook~\cite{fan2020beyond}. You can see an example of this attack in table~\ref{tab:table100}.
\begin{algorithm}[!h]
    \caption{Word-level attack}
    \begin{algorithmic}
        \Require{Machine translation model $T$}
        \Function{GetCandidates}{$\mathcal{M}$, x, y, token\_id}
            \ind\If{x[token\_id] in $T[L]$}
                    \\
                    \ind\ind tokens = $T[L]$[x[token\_id]]\\
                    \ind\ind x[token\_id] = tokens\\
                    \ind\ind y[token\_id] = ExtendSlotLabels(y[token\_id], len(tokens))
            \EndIf \\
            \ind\Return x, y
        \EndFunction
    \end{algorithmic}\label{alg:algorithm1}
\end{algorithm}
\begin{table}[H]
    \centering
    \caption{Example of attacking XLM-RoBERTa (xlm-r) with word-level attack.}
	\resizebox{0.9\textwidth}{!}{
		\begin{tabular}{>{\bfseries}l|ccccccccc}
			\toprule
			Utterance en &what & are & the & flights & from & las & vegas & to & ontario \\ \midrule
			Utterance adv &what & sind & die & flights & from & las & vegas & to & ontario \\ \bottomrule
		\end{tabular}
	}\label{tab:table100}
\end{table}
\subsubsection{Phrase-level adversaries} The second attack (Algorithm~\ref{alg:algorithm2}) generates equivalents from other languages by building alignments between sentences in different languages. One sentence is a translation of another; we utilize the fact that we have a parallel dataset. Candidates for each token are defined as tokens from the embedded sentence into which the token was aligned. For aligning sentences, we use the awesome-align model based on m-BERT~\cite{dou2021word}. You can see an example of this attack in table~\ref{tab:table101}.
\begin{algorithm}[!h]
    \caption{Phrase-level attack}
    \begin{algorithmic}
        \Require{Sentences alignment $A$}
        \Function{GetCandidates}{$\mathcal{M}$, x, y, token\_id}
            \If{x[token\_id] in $A[L]$}
                \\
                \ind\ind tokens = $A[L]$[x[token\_id]]\\
                \ind\ind x[token\_id] = tokens\\
                \ind\ind y[token\_id] = ExtendSlotLabels(y[token\_id], len(tokens))
            \EndIf \\
            \ind\Return x, y
        \EndFunction
    \end{algorithmic}\label{alg:algorithm2}
\end{algorithm}
\begin{table}[H]
    \centering
    \caption{Example of attacking XLM-RoBERTa (xlm-r) with phrase-level attack.}
	\resizebox{0.9\textwidth}{!}{
		\begin{tabular}{>{\bfseries}l|ccccccccc}
			\toprule
			Utterance en &please & find & flights & available & from & kansas & city & to & newark \\ \midrule
			Utterance adv &encontre & find & flights & disponíveis & from & kansas & city & para & newark \\ \bottomrule
		\end{tabular}
	}\label{tab:table101}
\end{table}
\subsection{Adversarial pre-training method protects from adversarial attacks} Adversarial pre-training protects the model against the proposed adversarial attacks. It most likely allows the model to increase the performance not only at adversarial perturbations but also on real data with code-switching. However, this is only a hypothesis since there is no real-life code-switched ToD data. 

The adversarial pre-training method relies on domain adaptation techniques and has several steps:
\begin{enumerate}
    \item Generating adversarial training set for masked language modeling task.
    \item Fine-tuning language model's body on the new generated set in masked language modeling task.
    \item Loading fine-tuned model's body before training for joint intent classification and slot labeling task.
\end{enumerate}

\subsubsection{Generating adversarial training set} To generate an adversarial training set, we use an adaptation of the phrase-level algorithm of the adversarial attack (Algorithm~\ref{alg:algorithm4}). The difference is that tokens are replaced with their equivalents with the probability of $0.5$. Thus, a trained model is not required to generate the sample. The adversarial training set is a concatenation of generated sets for all languages in the dataset except English. Each subset is generated by embedding the target language into the training set of the MultiAtis++ dataset in English. After generation, we get six subsets of 4884 sentences each. The final adversarial training set consists of 29304 sentences; we divide it into training and test sets in a ratio of 9 to 1.
\begin{algorithm}[!h]
    \caption{Generating adversarial training set}
    \begin{algorithmic}
        \Require{Training dataset $X$, set of embedded languages $L_1, \dots L_n$}
        \Ensure{Adversarial training set $X'$} \\
        X' = [~]
        \For{$L$ in $L_1, \dots L_n$}
            \For{x in X}
                \For{i in permutation(len(x))}
                    \\
                    \ind\ind\ind Candidates = GetCandidates($\mathcal{M}$, x, y, token\_id = i)
                    \ind\If{Candidates and $\mathcal{U}$(0, 1) > 0.5}
                            \\
                            \ind\ind\ind\ind x, \_ = random.choice(Candidates)
                    \EndIf
                \EndFor \\
                \ind\ind X`.append(x)
            \EndFor
        \EndFor \\
        \Return X'
    \end{algorithmic}\label{alg:algorithm4}
\end{algorithm}
\subsubsection{Fine-tuning model's body} After generating the adversarial training set, we fine-tune the pre-trained multilingual model. The model is trained with the masked language modeling objective~\cite{devlin-etal-2019-bert}. We select 15\% of tokens and predict them using the model to train a model for such a task. 80\% of the selected tokens are replaced with the mask token, 10\% are replaced with random words from the model's dictionary, the remaining 10\% remain unchanged. After fine-tuning, we dump the body of the model for future use.

\subsubsection{Loading fine-tuned model's body} Before training the multilingual model for the task of join intent classification and slot labeling, we load the fine-tuned body of the model. For models that have been pre-trained using the adversarial pre-training method, we will add the suffix \textbf{adv} to the name.

\section{Experimental results} We will compare models trained only on the English training set (zero-shot models) and the whole training set (full models). We will evaluate the quality by three metrics - accuracy for intents, f1-score for slots and semantic accuracy. We found that zero-shot models have significantly worse quality than full ones, not only in languages other than English but even in English.

\subsubsection{Joint intent classification and slot labelling} 
\begin{figure}[h]
    \centering
    \includegraphics[width=0.95\textwidth]{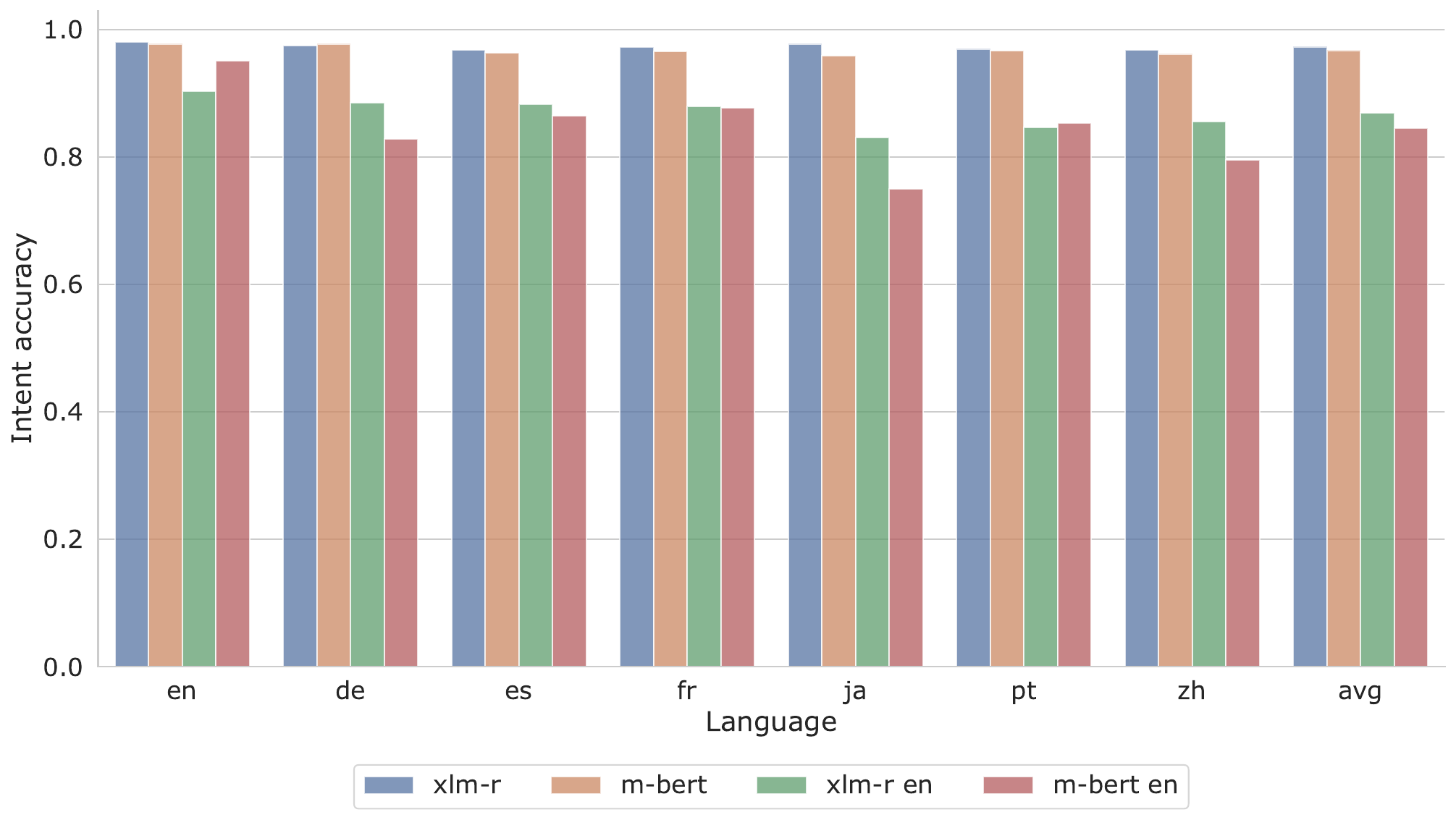}
    \caption{Model comparison by \textbf{Intent accuracy} metric.}\label{fig:figure100}
\end{figure}

We have achieved strong performance for the problem of classifying intents and filling in slots. On the test sample, full models showed an average of 97\% correct answers for intents, and zero-shot ones, on average, 85\%. Full models showed 0.93 f1-score for slots, zero-shot 0.68. Full models showed 79\% of completely correctly classified sentences and zero-shot ones - about 26\%. This shows that zero-shot learning is not capable of competing with full learning in this particular task.
In the Fig~\ref{fig:figure100}, you can see the comparison between models and languages by Intent accuracy metric. The additional results are provided in the project's repository.
\subsubsection{Attacking models} 
\begin{figure}[H]
    \centering
    \includegraphics[width=0.95\textwidth]{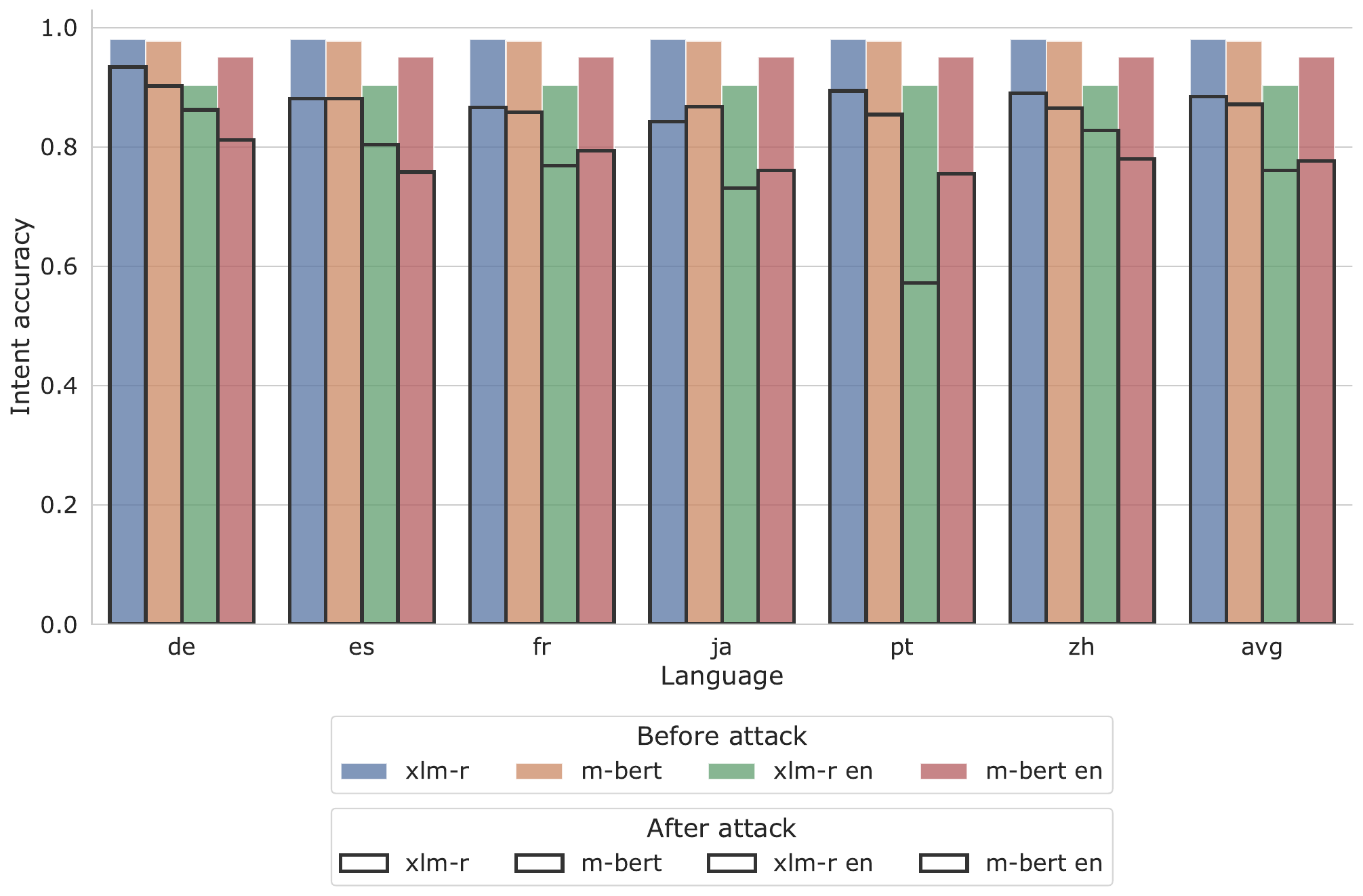}
    \caption{Model comparison after \textbf{word-level} attack by \textbf{Intent accuracy} metric.}\label{fig:figure101}
\end{figure}

We attacked all the models using our two algorithms. We found that the word-level attack turned out to be more advanced, leading to lower performance. For full models, the quality of intents fell from 98\% to 88\%, for zero-shot models from 92\% to 77\%. For full models, the quality by slots fell from 0.95 to 0.6, for zero-shot models from 0.88 to 0.48. For full models, the proportion of entirely correctly classified sentences fell from 83\% to 14\%, for zero-shot ones from 60\% to 5\%. Figure~\ref{fig:figure101} compares results before and after the word-level attack with the Intent accuracy metric.

\begin{figure}[h]
    \centering
    \includegraphics[width=0.95\textwidth]{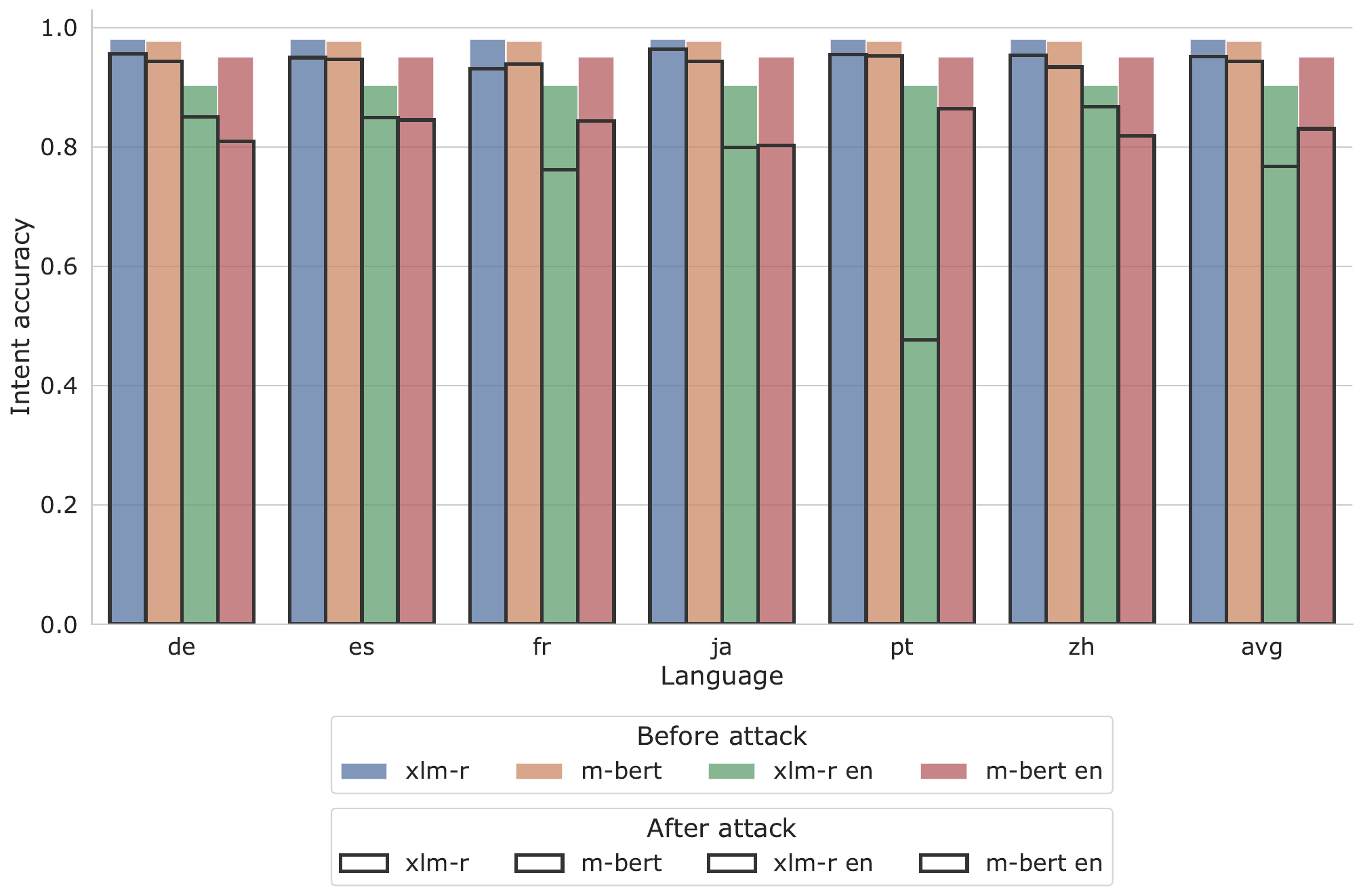}
    \caption{Model comparison after \textbf{phrase-level} attack by \textbf{Intent accuracy} metric.}\label{fig:figure102}
\end{figure}

We also got that the phrase-level attack turned out to be softer and gave a higher quality compared to the word-level attack. For full models, the quality of intents fell from 98\% to 95\%, for zero-shot models from 92\% to 80\%. For full models, the quality by slots fell from 0.95 to 0.7, for zero-shot models from 0.88 to 0.55. For full models, the proportion of entirely correctly classified sentences fell from 83\% to 35\%, for zero-shot ones from 60\% to 10\%. In the Fig~\ref{fig:figure102}, you can see the quality comparison after the phrase-level attack for the Intent accuracy metric.

\subsubsection{Adversarial pre-training} To protect the models from attacks, we fine-tuned the bodies for both models and loaded them before training for the task of joint intent classification and slot filling. We found that the defense had almost no effect on the full models in terms of quality on the test set. For zero-shot models, the effect on the test set is ambiguous - the quality by intents fell for Asian languages but increased slightly for all others. As for the slots, we observe a negative effect for the m-BERT model and a positive effect for the XLM-RoBERTa model.

For the word-level attack, a slight deterioration in the quality of intents for Asian languages is noticeable, and a positive effect for other languages. After the adversarial pre-training, the quality of slots increased for all models, which ultimately results in an almost two-fold increase in the proportion of entirely correctly classified sentences for zero-shot models and about 15\% relative improvement for full models.
\begin{figure}[H]
    \centering
    \includegraphics[width=0.95\textwidth]{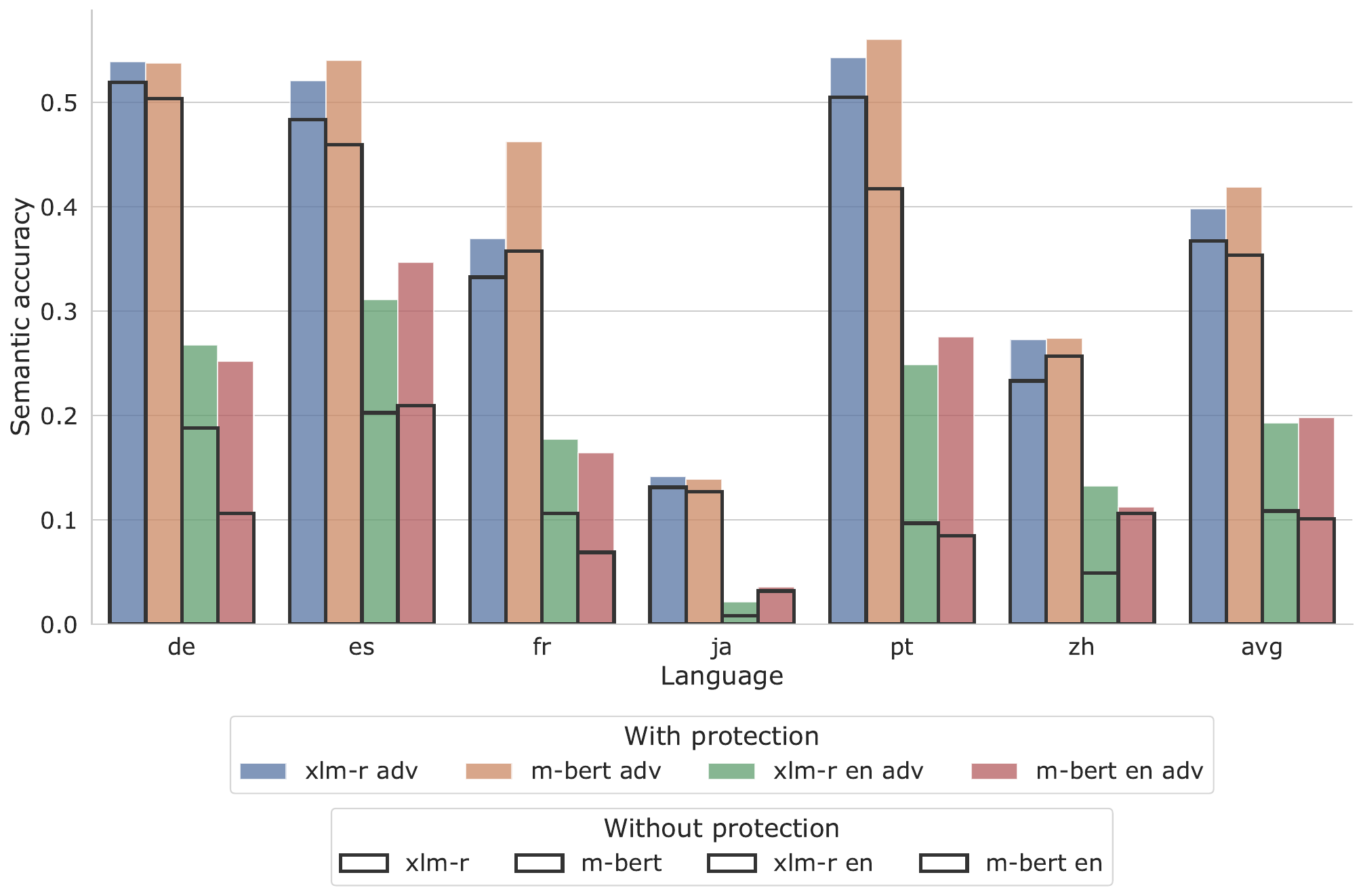}
    \caption{Model comparison \textbf{with protection} after \textbf{phrase-level} attack by \textbf{Semantic accuracy} metric.}\label{fig:figure103}
\end{figure}
Again, for the phrase-level attack, a slight deterioration in the quality of intents for Asian languages is noticeable, and a positive effect for other languages. After the defense, the quality of slots dropped slightly for Asian languages and increased significantly for the rest. This results in a two-fold increase in the proportion of entirely correctly classified sentences for zero-shot models and about 15\% relative improvement for full models (Fig~\ref{fig:figure103}).

\section{Discussion} We approached the problem of recognizing intents and filling in slots for a multi-lingual ToD system. We study the effect of switching codes on two multi-lingual language models, XLM-RoBERTa and m-BERT. We showed that switching codes could become a noticeable problem when applying language models in practice using two gray-box attacks. However, the defense method shows promising results and helps to improve the quality after the model is attacked. 

%As future work, we consider the analysis of other multilingual models, the construction of new more realistic attacks that simulate code-switching and the search for new defense algorithms against such attacks.

\section{Conclusion} This paper presents an adversarial attack on multilingual task-oriented dialog (ToD) systems that simulates code-switching. This work is motivated by research and practical needs. First, the proposed attack reveals that pre-trained language models are vulnerable to synthetic code-switching. To this end, we develop a simplistic defense technique against code-switched adversaries. Second, our work is motivated by the practical needs of multilingual ToDs to cope with code-switching, which is seen as an essential phenomenon in multicultural societies. Future work directions include evaluating how plausible and naturally-sounding code-switched adversaries are and adopting similar approaches to model-independent black-box scenarios.

\section{Acknowledgements}
The article was prepared within the framework of the HSE University Basic Research Program.

\bibliographystyle{splncs04}
\bibliography{papers}

\end{document}